\documentclass[10pt,twocolumn,letterpaper]{article}

\usepackage{cvpr}              

\usepackage{graphicx}
\usepackage{amsmath}
\usepackage{amssymb}
\usepackage{booktabs}
\usepackage{multirow}
\usepackage{makecell}
\usepackage{amssymb}
\usepackage{bbm}
\usepackage{color}
\usepackage{bbding}
\usepackage[table]{xcolor}
\usepackage{threeparttable}

\usepackage[pagebackref,breaklinks,colorlinks]{hyperref}
\usepackage{stmaryrd}
\usepackage{amssymb}

\newcommand{\keypoint}[1]{\vspace{0.1cm}\noindent\textbf{#1}}

\usepackage[capitalize]{cleveref}
\crefname{section}{Sec.}{Secs.}
\Crefname{section}{Section}{Sections}
\Crefname{table}{Table}{Tables}
\crefname{table}{Tab.}{Tabs.}

\def\cvprPaperID{1304} 
\def\confName{ICCV}
\def\confYear{2023}

\begin{document}

\title{DETA: Denoised Task Adaptation for Few-Shot Learning}

\author{Ji Zhang\textsuperscript{1}\hspace{-0.2cm}\and
Lianli Gao\textsuperscript{2}\hspace{-0.0cm}\thanks{Corresponding author. \textit{This work is published at}  \href{https://openaccess.thecvf.com/content/ICCV2023/html/Zhang_DETA_Denoised_Task_Adaptation_for_Few-Shot_Learning_ICCV_2023_paper.html}{\textit{ICCV 2023}}.}\and
Xu Luo\textsuperscript{1}\hspace{-0.2cm}\and
Hengtao Shen\textsuperscript{1}\hspace{-0.2cm}\and
Jingkuan Song\textsuperscript{2} \and 
\textsuperscript{1}University of Electronic Science and Technology of China (UESTC) \\ \textsuperscript{2}Shenzhen Institute for Advanced Study, UESTC \\
{\tt\small \{jizhang.jim,juana.alian\}@gmail.com}
}



\maketitle

\begin{abstract}
   Test-time task adaptation in few-shot learning aims to adapt a pre-trained task-agnostic model for capturing task-specific knowledge of the test task, relying only on few-labeled support samples. 
   Previous approaches generally focus on developing advanced algorithms to achieve the goal, while neglecting the inherent problems of the given support samples. 
   In fact, with only a handful of samples available, the adverse effect of either the image noise (a.k.a. $\mathtt{X}$-noise) or the label noise (a.k.a. $\mathtt{Y}$-noise) from support samples can be severely amplified. 
   To address this challenge, in this work we propose \textbf{DE}noised \textbf{T}ask \textbf{A}daptation (\textbf{DETA}), a first, unified image- and label-denoising framework orthogonal to existing task adaptation approaches.
   Without extra supervision, DETA filters out task-irrelevant, noisy representations by taking advantage of both global visual information and local region details of support samples. 
    On the challenging Meta-Dataset, DETA consistently improves the performance of a broad spectrum of baseline methods applied on various pre-trained models.
   Notably, by tackling the overlooked image noise in Meta-Dataset, DETA establishes new state-of-the-art results.
   Code is released at \url{https://github.com/JimZAI/DETA}.

\end{abstract}

\section{Introduction}
\label{s1}
Few-Shot Learning (FSL) refers to rapidly deriving new knowledge from a limited number of samples, a central capability that humans naturally possess, but “data-hungry” machines still lack. 
Over the past years, a community-wide enthusiasm has been ignited to narrow this gap,
especially in fields such as computer vision\cite{snell2017prototypical,li2017meta,he2022masked}, machine translation \cite{xing2019adaptive,bragg2021flex,liu2019few} and reinforcement learning \cite{finn2017model,hong2021reinforced,nichol2018first}.
\begin{figure}
\setlength{\abovecaptionskip}{0.15cm}  
\setlength{\belowcaptionskip}{-0.25cm} 
		\centering 
		\includegraphics[width=0.9\linewidth]{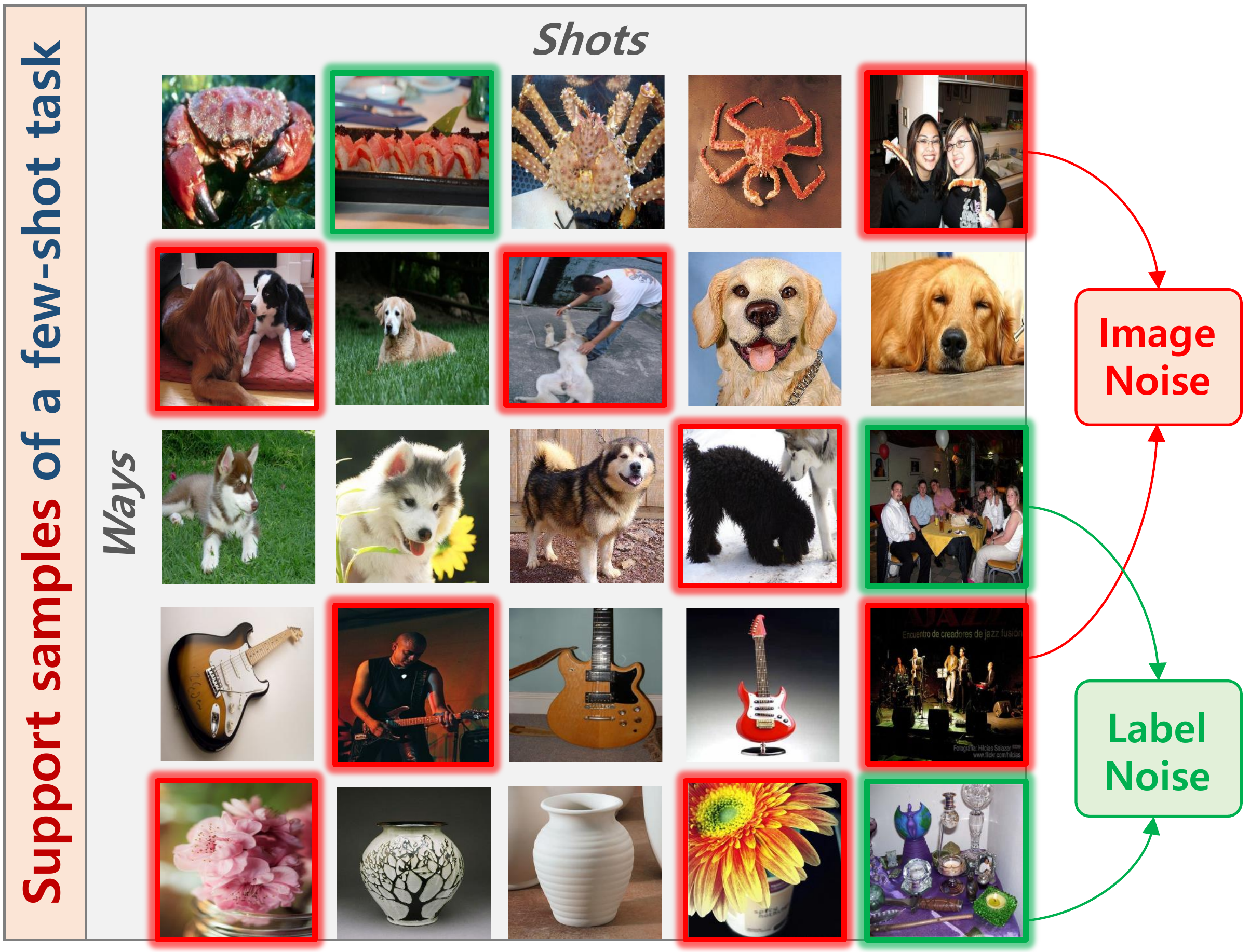}
		\caption{Dual noises in the support samples of a few-shot task. \textbf{Image noise} (a.k.a. $\mathtt{X}$-noise): the target object regions are often obscured by interfering factors such as cluttered backgrounds, image corruption, etc. \textbf{Label noise} (a.k.a. $\mathtt{Y}$-noise): mislabeled samples. 
		The goal of this work is to develop a first, unified image- and label-denoising framework for reliable task adaptation.
		}
		\label{f1}
\end{figure}

The general formulation of FSL involves two stages: 
\textbf{1)} \textit{training-time} task-agnostic knowledge accumulation, and \textbf{2)} \textit{test-time} task-specific knowledge acquisition, a.k.a. task adaptation.
In particular, the former stage seeks to pre-train a task-agnostic model on large amounts of training samples collected from a set of \textit{base} classes. 
While the latter targets adapting the pre-trained model for capturing task-specific knowledge of the few-shot (or test) task with \textit{novel} classes, given a tiny set of labeled support samples. 
Early progress in FSL has been predominantly achieved using the idea of meta-learning, which aligns the learning objectives of the two stages to better generalize the accumulated knowledge towards few-shot tasks \cite{snell2017prototypical,xing2019adaptive,nichol2018first}.
Nevertheless, recent studies \cite{luo2023closer,requeima2019fast,li2022cross,gidaris2019generating}  revealed that a good test-time task adaptation approach with any pre-trained models -- no matter what training paradigms they were learned by, can be more effective than sophisticated meta-learning algorithms.
Furthermore, with the recent success in model pre-training techniques  \cite{he2020momentum,han2021transformer,liu2021swin}, designing efficient {adapter-based} \cite{li2022cross,xuexploring,li2021universal} or {finetuning-based} \cite{asimplebaseline,hu2022pushing} task adaptation algorithms that can flexibly borrow “free” knowledge from a wide range of pre-trained models is therefore of great practical value, and has made remarkable progress in FSL.

\begin{figure} 
\setlength{\abovecaptionskip}{0.13cm}  
\setlength{\belowcaptionskip}{-0.3cm} 
		\centering
		\includegraphics[width=1\linewidth]{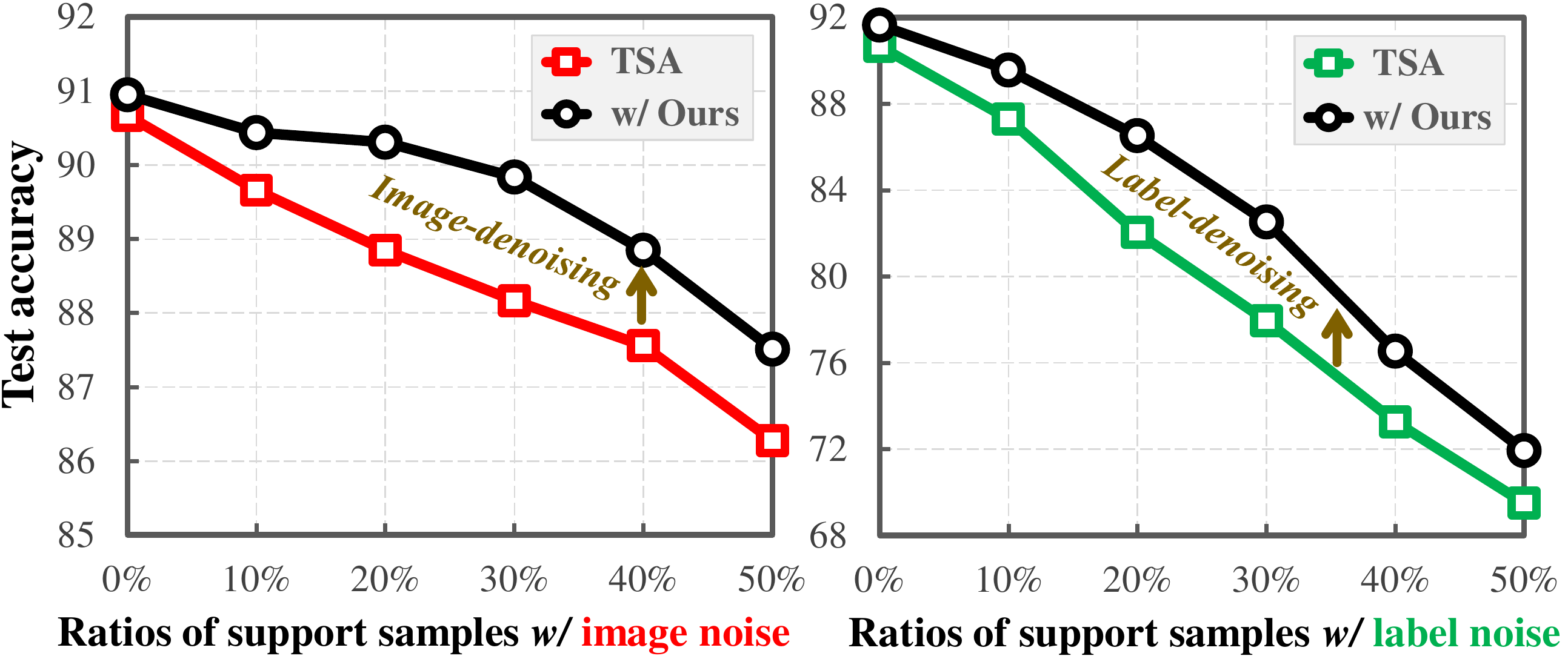}
		\caption{Quantitative evidence that image- or label-noisy support samples degrades test-time task adaptation performance.
		The results are averaged over 100 5-way 10-shot tasks sampled from the five classes in Figure \ref{f1}. 
		\textit{Image-noisy} samples here are manually selected from all samples of the five classes. \textit{Label-noisy} samples for each class are generated by uniformly changing the label to that of other four classes.
		The baseline scheme TSA \cite{li2022cross} is applied to a RN-18 pre-trained on ImNet-MD \cite{triantafillou2019meta} for task adaptation. As seen, the dual noises negatively impact task adaptation performance, and our method consistently improves the baseline under various ratios of image- or label-noisy support samples.}
		\label{f2}  
\end{figure}

Despite the encouraging progress, existing approaches mostly focus on developing advanced algorithms to mine task-specific knowledge for few-shot tasks, while neglecting the inherent problems of the given support samples.
Unfortunately, the set of support samples collected from the open world, no matter how small, can be unavoidably polluted by noises. 
As illustrated in Figure \ref{f1}, either the image noise (a.k.a. $\mathtt{X}$-noise) or the label noise (a.k.a. $\mathtt{Y}$-noise) could arise at possibly every phase of the task lifecycle\footnote{In more challenging FSL scenarios, some or even all of the few examples are collected by an agent from a dynamic environment rather than relying on humans, the dual noises become more common in this context.}.
It has been well recognized that a tiny portion of image-noisy \cite{luo2021rectifying,kong2021resolving} or label-noisy \cite{liu2015classification,song2022learning} samples can compromise the model performance to a large extent.
When it comes to test-time task adaptation, the adverse effects of the dual noises can be remarkably magnified owing to the \textit{scarcity} of support samples, as quantitatively proven in Figure \ref{f2}.
Despite being harmful and inevitable, as far as we know, both image noise and label noise have received considerably less attention in test-time task adaptation.
\textbf{{This begs the following questions:}}
\textbf{1)} Is it possible to design a method to tackle the two issues in a unified framework?
\textbf{2)} Whether the designed method can be orthogonal to existing task adaptation approaches, so as to achieve robust FSL?

\begin{figure*}[ht]
\setlength{\abovecaptionskip}{0.15cm}  
\setlength{\belowcaptionskip}{-0.2cm} 
		\centering
		\includegraphics[width=1\linewidth]{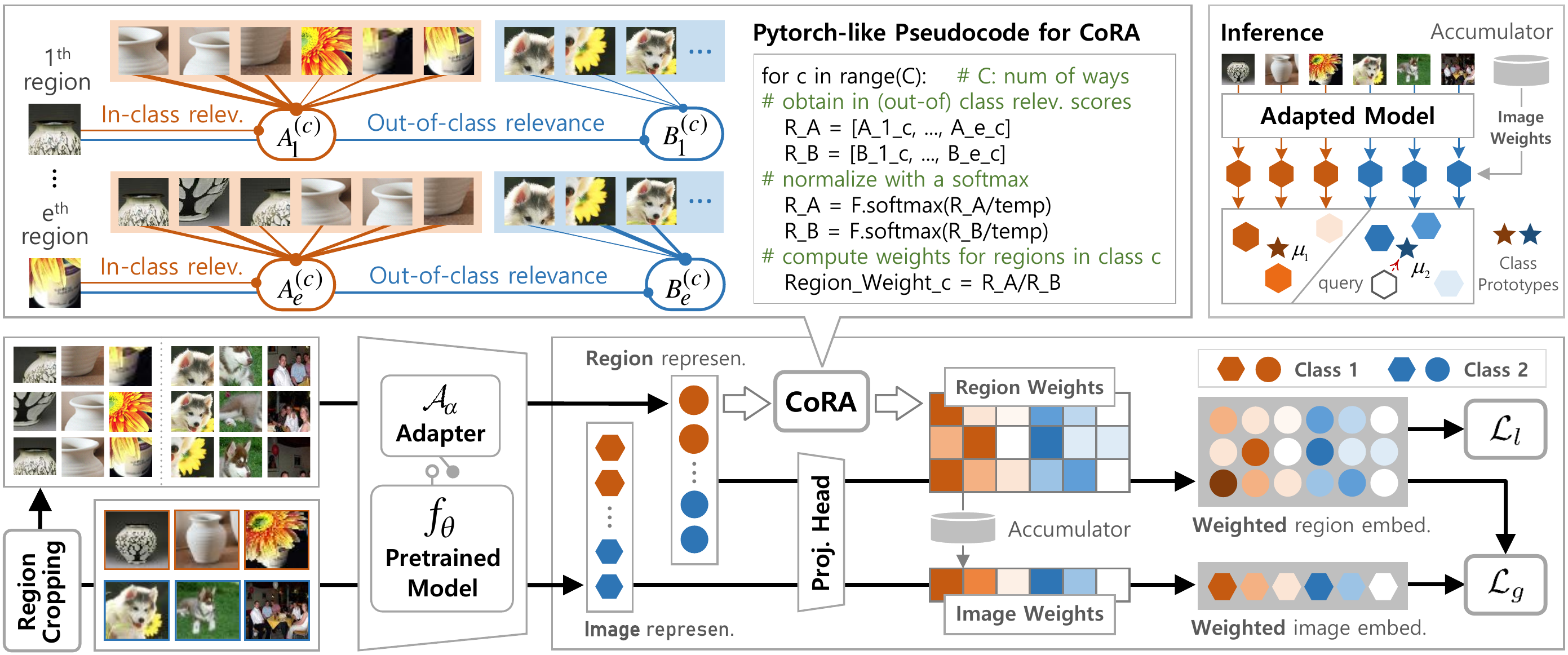} 
		\caption{An overview of the proposed \textbf{DETA} (in a 2-way 3-shot exemple). 
        During each iteration of task adaptation, the images together with a set of randomly cropped local regions of support samples are first fed into a pre-trained model $f_{{{\theta}}}$ (w/ or w/o a model-specific adapter $\mathcal{A}_{{{\alpha}}}$) to extract image and region representations.
		Next, a \textbf{Contrastive Relevance Aggregation(CoRA)} module takes the region representations as input to determine the weight of each region, based on which we can refine the image weights by a momentum accumulator.
		Finally, a \textbf{Local Compactness} loss $\mathcal{L}_l$, along with a \textbf{Global Dispersion} loss $\mathcal{L}_g$ are devised in a {weighted} embedding space for promoting the mining of task-specific (or clean) representations.
		At inference, we only retain the adapted model $f_{{{\theta}^*}}$ (or $f_{[{\theta};{\alpha}^*]}$) to produce image representations of support samples, on which we can build a noise-robust classifier guided by the refined image weights in the accumulator.
		} \label{pip}
\end{figure*}

In this work, we answer the above questions by proposing {{\textbf{DE}noised  \textbf{T}ask \textbf{A}daptation (\textbf{DETA})}}, a first, unified image- and label-denoising framework for FSL.
The key idea of DETA is to simultaneously filter out task-irrelevant (i.e. noisy) local region representations of \textit{image-noisy} samples, as well as global image representations of \textit{label-noisy} samples, relying only on the interrelation among the given support samples of few-shot tasks.
To this end, a parameter-free {\textit{contrastive relevance aggregation (CoRA)}} module is first designed to determine the weights of regions and images in support samples, based on which two losses are proposed for noise-robust (or reliable) task adaptation: 
a \textit{local compactness} loss $\mathcal{L}_l$ that promotes the intra-class compactness of \textit{clean} regions, along with a \textit{global dispersion} loss $\mathcal{L}_g$ that encourages the inter-class dispersion of \textit{clean}, image-level class prototypes. 
The two losses complement each other to take advantage of both global visual information and local region details of support samples to softly ignore the dual noises during the optimization.
An overview of our DETA framework is shown in Figure \ref{pip}.

\keypoint{Flexibility and Strong Performance.} 
The proposed DETA is orthogonal to existing \textit{adapter}-based task adaptation ($\mathtt{A}$-$\mathtt{TA}$) and \textit{finetuning}-based task adaptation ($\mathtt{F}$-$\mathtt{TA}$) paradigms, therefore can be plugged into any types of these approaches to improve model robustness under the joint (image, label)-noise.
On average, by performing image-denoising on the vanilla Meta-Dataset (MD) \cite{triantafillou2019meta},
DETA improves the classification accuracy of $\mathtt{A}$-$\mathtt{TA}$, $\mathtt{F}$-$\mathtt{TA}$ baselines by \textbf{1.8}\%$\sim$\textbf{1.9}\%, \textbf{2.2}\%$\sim$\textbf{4.1}\%, respectively (Table \ref{t1}). 
In particular, by tackling the overlooked image noise in the vanilla MD, DETA further boosts the state-of-the-art TSA \cite{li2022cross} by \textbf{1.8}\%$\sim$\textbf{2.1}\% (Table \ref{sota}).
Also, by conducting label-denoising on the label-corrupted MD, DETA outperforms $\mathtt{A}$-$\mathtt{TA}$, $\mathtt{F}$-$\mathtt{TA}$ baselines by \textbf{1.8}\%$\sim$\textbf{4.2}\%, \textbf{2.8}\%$\sim$\textbf{6.1}\%, respectively (Table \ref{table2}).

\keypoint{Contributions.} 
To summarize, our contributions are threefold.
\textbf{1)} 
We propose DETA, a first, unified image- and label-denoising framework for FSL. 
\textbf{2)} Our DETA can be flexibly plugged into both adapter-based and finetuning-based task adaptation paradigms.
\textbf{3)} Extensive experiments on  Meta-Dataset show the effectiveness and flexibility of DETA.

\section{Related Work}
\label{sec:related work}
\keypoint{Few-shot Learning.} 
Generalizing from a limited amount of samples 
has been proven challenging for most existing deep learning models. 
Prevalent FSL approaches learn new concepts under scarce supervision by a meta-learning setting \cite{oreshkin2018tadam,liang2021boosting,ye2020few,hou2019cross,fu2022generalized,shao2023fads,zhang2022progressive,Fu_2023_CVPR,zhang2022free,zhang2021curriculum}.
In \textbf{{Sup. Mat. (E.1)}}, we present a review of the literature on FSL approaches. 

\keypoint{Test-time Task Adaptation in FSL.} 
Recent progress revealed that {when there exists severe \textit{category}/\textit{domain shift}  between base classes and few-shot tasks, without performing test-time task adaptation, the generalization of any pre-trained models would decrease remarkably}\cite{requeima2019fast,bateni2020improved,luo2023closer}. 
Various attempts have been made to adapt the pre-trained models to few-shot tasks by devising model-specific adapters, \textit{e.g.}, the residual adapter TSA \cite{li2022cross} for ResNets \cite{he2022masked}, the self-attention adapter eTT \cite{xuexploring} for ViTs \cite{dosovitskiy2020image}. 
A survey of test-time task adaptation is presented in \textbf{{Sup. Mat. (E.2)}}.

\keypoint{Data-denoising for FSL.}
The training data collected from the open world are unavoidably polluted by image noise or label noise, which may compromise the performance of the learned models \cite{arpit2017closer,song2022learning,iscen2020graph}.
Limited works in FSL considered the influence of image noise ~\cite{luo2021rectifying,xu2022alleviating} or label noise ~\cite{liang2022few,mazumder2021rnnp} on model generalization. 
Additionally, they mainly focus on dealing with noises in base classes rather than in the few-shot task.
Particularly, Liang et al.~\cite{liang2022few} for the first time explored the label noise problem in FSL.
Differences between the work \cite{liang2022few} and ours are threefold. 
\textbf{1)} We aim to address both the image and label noises from support samples, where every sample is of great value in characterizing the few-shot task.
\textbf{2)} We take advantage of both global visual information and local region details to achieve the goal.
\textbf{3)} Our method is orthogonal to both adapter-based and finetuning-based
task adaptation methods.
Even so, Liang et al.~\cite{liang2022few} do bring a lot of inspiration to our method.

\keypoint{Cross-image Alignment for Representation Learning.}
A plethora of cross-image alignment based FSL methods have recently been developed to extract more discriminative representations \cite{zhang2020deepemd,hou2019cross,li2019revisiting,wu2019parn,wertheimer2021few,Mo_ICML_2023}.
Those methods highlight important local regions by aligning the local features between the support and query samples of few-shot tasks.
Despite the impressive performance, those \textit{none-adaptation} methods are unable to capture task-specific representations when there exists severe  \textit{category shift} or \textit{domain shift}  between base classes and few-shot tasks \cite{luo2023closer,hu2022pushing}.
Moreover, we often overlook the fact that owing to the small sample size in few-shot tasks, negligible computational cost is required to model the relationships of the support samples.

\section{Methodology}
In this section, we elaborate our proposed DETA. Before that, we introduce some preliminary concepts
about test-time task adaptation in FSL, and the mechanism of adapter-based or finetuning-based task adaptation.

\subsection{Preliminary}
\label{pre}
Assume we have a pre-trained task-agnostic model $f_\theta$ parameterized by ${{\theta}}$, which serves as a feature backbone to output a \textit{d}-dimensional representation for each input image.
Test-time task adaptation seeks to adapt $f_{\theta}$ to the test task $T=\{S,Q\}$, by deriving task-specific knowledge on the few-labeled support samples ${S}=\{(\boldsymbol{x}_i,y_i)\}_{i=1}^{N_s}$, consisting of $N_s$ {\textit{image}}-{{\textit{label}}} pairs from $C$ novel classes, \textit{i.e.}, $y_i \in \{1,..., C\}$.
It is expected that the adapted model can correctly partition the $N_q$ query samples ${{Q}}=\{(\boldsymbol{x}_i)\}_{i=1}^{N_q}$ to the $C$ classes in the \textit{representation} space. 
If there are exactly $K$ support samples in each of these $C$ classes, the task is also called a $C$-way $K$-shot task.

\keypoint{Adapter-based Task Adaptation ($\mathtt{A}$-$\mathtt{TA}$)}. 
The goal of $\mathtt{A}$-$\mathtt{TA}$ is to capture the knowledge of the test task by attaching a model-specific adapter $\mathcal{A}_{{{\alpha}}}$ parameterized by ${{\alpha}}$ to the pre-trained model $f_{{{\theta}}}$. 
During task adaptation, the parameters of $f_{{{\theta}}}$, ${\theta}$, are frozen and only the parameters ${\alpha}$ are optimized from scratch using the support samples:
\begin{equation} 
\alpha := \alpha - \gamma \nabla_\alpha \mathcal{L}^{{S}}\big([{\theta};{\alpha}]\big),
\end{equation}
where $\gamma$ is the learning rate, and
\begin{equation} 
\mathcal{L}^{{S}}([{\theta};{\alpha}])=\frac{1}{N_s}\sum_{(\boldsymbol{x},y) \in {S}}\ell\big(h(f_{[{\theta};{\alpha}]}(\boldsymbol{x});S), y\big),
\end{equation}
where $\ell$ is cross-entropy loss, $f_{[{\theta};{\alpha}]}$ indicates the feature backbone appended with the adapter, $h$ is a non-parametric classifier head capable of producing a softmax probability vector whose dimensionality equals $C$.
Notably, the recent $\mathtt{A}$-$\mathtt{TA}$ scheme TSA \cite{li2022cross} achieved state-of-the-art results on Meta-Dataset \cite{triantafillou2019meta}, by integrating a residual-adapter into the pre-trained URL \cite{li2021universal} model (w/ RN-18), and setting $h$ to the nonparametric Nearest Centroid Classifier (NCC) \cite{mensink2013distance}. 

\keypoint{Finetuning-based Task Adaptation ($\mathtt{F}$-$\mathtt{TA}$)}. 
$\mathtt{A}$-$\mathtt{TA}$ requires model-specific adapters to adapt different pre-trained models,  {e.g.,} the residual adapter TSA \cite{li2022cross} for ResNets \cite{he2022masked}, the self-attention adapter eTT \cite{xuexploring} for ViTs \cite{dosovitskiy2020image}. 
In contrast, $\mathtt{F}$-$\mathtt{TA}$, originated from transfer learning literature~\cite{kornblith2019better} and introduced into FSL by MAML~\cite{finn2017model}, directly finetunes the parameters $\theta$ of any pre-trained model $f_{{{\theta}}}$ at test time, {i.e.}, $\theta := \theta - \gamma \nabla \mathcal{L}^{{S}}(\theta)$, and is thus model-agnostic.

\subsection{Overview}
Our framework DETA is illustrated in Figure \ref{pip}, which mainly consists of the following steps. 
\textbf{For each iteration:}

\keypoint{Step-1.} A feature backbone $f$ takes the images and a set of randomly cropped image regions of the support samples as input to obtain image and region representations. 

\keypoint{Step-2.} A\textit{ contrastive relevance aggregation} (CoRA) module takes the region representations as input to calculate the weights of different regions, based on which we can determine the image weights by a momentum accumulator.

\keypoint{Step-3.} A projection head maps the high-dimensional image and region representations to a lower dimensional embedding space, where a \textit{local compactness} loss $\mathcal{L}_l$ and a \textit{global dispersion} loss $\mathcal{L}_g$ are developed on the weighted region and image embeddings to promote the mining of task-specific knowledge from support samples.

\keypoint{Step-4.} The calculated $\mathcal{L}_l$ and $\mathcal{L}_g$ are jointly used to update the parameters of the projection head and the feature backbone $f$, i.e., $\alpha$ in $f_{[{\theta};{\alpha}]}$ for $\mathtt{A}$-$\mathtt{TA}$, $\theta$ in $f_{\theta}$ for $\mathtt{F}$-$\mathtt{TA}$.

\subsection{Contrastive Relevance Aggregation}
\label{cora}
The motivation of CoRA is that a region, which shows higher relevance (or similarity) to \textit{in-class} regions while lower relevance to \textit{out-of-class} regions, is more likely to be the object region and should be assigned a larger weight.

Given the support samples ${S}=\{(\boldsymbol{x}_i,y_i)\}_{i=1}^{N_s}$ of a test-time task, we first randomly crop $k$ local regions of size $M\times M$ for every image $\boldsymbol{x}_i$.
Next, the original image together with all cropped regions of each sample are fed into $f$ to generate image representation $\boldsymbol{{{z}}}_i$ and region representations $Z_i =\{\boldsymbol{{{z}}}_{ij}\}_{j=1}^{k}$. 
Let $Z^{(c)}=\bigcup_{i=1}^{N_{c}}Z_i$ denote the collection of representations of cropped regions in class $c$, $\textbf{\textit{Z}}=\bigcup_{c=1}^{C}Z^{(c)}$ the set of all representations of cropped regions, where $N_c$ is the number of images in class $c$. 
For each region representation $\boldsymbol{{{z}}}_{ij}$ in $Z_i$, we construct its \textit{in-class} and \textit{out-of-class} region representation sets as 
${I}(\boldsymbol{z}_{ij})=Z^{(c)}\setminus Z_i$ and ${O}(\boldsymbol{z}_{ij})=\textbf{\textit{Z}}\setminus Z^{(c)}$, respectively.
Note that in $I(\boldsymbol{z}_{ij})$, the other $k-1$ intra-image representations are dropped to alleviate their dominating impacts. 
CoRA calculates the weight of each region based on the global statistics of in-class and out-of-class relevance scores, respectively formulated as
\begin{equation}
        \phi(\boldsymbol{z}_{ij}) = \frac{1}{|I(\boldsymbol{z}_{ij})|} \sum_{\boldsymbol{{{z}}}' \in I(\boldsymbol{z}_{ij})} \zeta(\boldsymbol{{{z}}}_{ij}, \boldsymbol{{{z}}}'),
        \label{eq1}
\end{equation}
\begin{equation}
        \psi(\boldsymbol{z}_{ij})
         = \frac{1}{|O(\boldsymbol{z}_{ij})|} \sum_{\boldsymbol{{{z}}}' \in O(\boldsymbol{z}_{ij})} \zeta(\boldsymbol{{{z}}}_{ij}, \boldsymbol{{{z}}}'),
         \label{eq2}
\end{equation}
\noindent where $\zeta(\cdot)$ indicates cosine similarity.
These scores are then normalized inside each class:
\begin{equation}
    \widetilde{\phi}(\boldsymbol{z}_{ij})\!=\!\frac{e^{{\phi(\boldsymbol{z}_{ij})}}}{\sum_{\substack{\boldsymbol{{{z}}}' \in Z^{(c)}}}{\!e^{\phi(\boldsymbol{{{z}}}')}}}, \, \widetilde{\psi}(\boldsymbol{z}_{ij})\!=\!\frac{e^{{\psi(\boldsymbol{z}_{ij})}}}{\sum_{\substack{\boldsymbol{{{z}}}' \in Z^{(c)}}}{\!e^{\psi(\boldsymbol{{{z}}}')}}}.
\end{equation}
Therefore, the final calculated region weight for $\boldsymbol{z}_{ij}$ can be defined as $\lambda_{ij}=\widetilde{\phi}(\boldsymbol{z}_{ij})/\widetilde{\psi}(\boldsymbol{z}_{ij})\in\mathbb{R}$. A pytorch-like pseudocode for CoRA is illustrated in Figure \ref{pip}.

\keypoint{A Momentum Accumulator for Image-weighting.}
Aside from weighting the local regions, we also need to assess the \textit{quality} of the images themselves for filtering out label-noisy samples.
Intuitively, the most direct way to determine the weight of an image $\boldsymbol{x}_i$, $\omega_i$, is to average the weights of all $k$ cropped regions belonging to it, {i.e.}, $\omega_i = \frac{1}{k} \sum_{j=1}^k \lambda_{ij}. $

However, the randomly cropped regions in different task adaptation iterations may have large variations, resulting the frailty of the calculated image weights.
A momentum accumulator is thus developed to cope with this issue by
\begin{equation}
    \omega^{t}_i =
\begin{cases} 
\frac{1}{k} \sum_{j=1}^k  \lambda_{ij}, & \mbox{if }t=1 \\
\gamma \omega^{t-1}_i+ \frac{1-\gamma}{k} \sum_{j=1}^k   \lambda_{ij}, & \mbox{if } t>1
\end{cases}
\end{equation}
where $\omega^{t}_i$ denotes the accumulated image weight of $\boldsymbol{x}_i$ in the $t$-th iteration of task adaptation, $\gamma$ is the momentum hyperparameter, and we set it to 0.7 in our method.
For brevity, we omit the superscript $t$ in the following sections.

\subsection{Noise-robust Task Adaptation}
DETA performs image- and label-denoising in a unified framework to achieve noise-robust task adaptation.  
To this end, DETA simultaneously 
\textbf{\textcircled{1}} \textbf{promotes the intra-class compactness of {\textit{clean}} regions} -- to filter out noisy local representations (e.g. cluttered backgrounds of image-noisy samples), and 
\textbf{\textcircled{2}} \textbf{encourages the inter-class dispersion of {\textit{clean}}, image-level class prototypes} -- to filter out noisy global representations (i.e. images of label-noisy samples).
To formalize our idea, we first map each image representation $\boldsymbol{{{z}}}_i$  and its region representations ${Z}_i=\{\boldsymbol{{{z}}}_{ij}\}_{j=1}^{k}$ to a low-dimensional embedding space by a projection head. 
The $l_2$ normalized image embedding and  $k$ region embeddings are denoted as $\boldsymbol{{{e}}}_i$ and  ${E}_i=\{\boldsymbol{e}_{ij}\}_{j=1}^{k}=\{\boldsymbol{r}_{\iota}\}_{\iota=1}^{k}$, respectively. 
Define $E^{(c)}$ and $\textbf{\textit{E}}$ similar to $Z^{(c)}$ and $\textbf{\textit{Z}}$.

\underline{\textbf{To achieve \textbf{\textcircled{1}}}}, we softly pull together (resp. push away) \textit{clean} regions from the same class (resp. different classes), guided by the calculated region weights of CoRA. 
For every pair of region embeddings $\boldsymbol{r}_i$ and  $\boldsymbol{r}_j$  from the same class and their region weights $\lambda_i$ and $\lambda_j$,  the loss function is
\begin{equation} 
    l(\boldsymbol{r}_i,\boldsymbol{r}_j) = - \log \frac{\exp( \lambda_i\boldsymbol{r}_i\cdot\lambda_j\boldsymbol{r}_j/\tau)}{\sum_{\boldsymbol{r}_v\in \textbf{\textit{E}}\setminus\boldsymbol{r}_i}\exp( \lambda_i\boldsymbol{r}_i\cdot\lambda_v\boldsymbol{r}_v/\tau)},
\end{equation}
where $\tau$ is a temperature parameter.
The objective function is equivalent to minimizing the following loss:
\begin{equation} 
        \mathcal{L}_{l} = \! \frac{1}{\sum_{c=1}^C\!\!\frac{kN_{c}\times(kN_{c}-1)}{2}}\!\sum_{c=1}^{C} \sum_{\boldsymbol{r}_i,\boldsymbol{r}_j\in E^{(c)}}\!\!\!\!\!\!\!\mathbbm{1}_{\boldsymbol{r}_i\ne\boldsymbol{r}_j}l(\boldsymbol{r}_i,\boldsymbol{r}_j). \,
\end{equation}
We term $\mathcal{L}_{l}$ \textit{local compactness} loss, since it encourages the intra-class compactness of clean local regions. 
By regularizing the task adaptation process with $\mathcal{L}_{l}$, task-irrelevant local representations from support samples (e.g. cluttered backgrounds of image-noisy samples) can be effectively filtered out during the optimization.

\underline{\textbf{To achieve \textbf{\textcircled{2}}}}, we propose a \textit{global dispersion} loss that encourages large distances among different class prototypes aggregated by clean images. 
Inspired by ProtoNet \cite{snell2017prototypical}, we assign region-level queries to image-level class prototypes in a soft manner, guided by the calculated image and region weights. 
Concretely, we first use all image embeddings $\{\boldsymbol{e}_{i}\}_{{}=1}^{N_s}$ to construct $C$ aggregated class prototypes as
    \begin{equation}
		\boldsymbol{\mu}_c = \frac{1}{N_c} \sum_{y_{i}=c} \omega_{i} \boldsymbol{e}_{i}, \,\,\, c = 1,2, ..., C,
	\end{equation}
where the impact of label-noisy samples from each class $c$ are weakened by a lower image weight $\omega_{i}$.
Next, we estimate the likelihood of every region embedding $\boldsymbol{r}_{j}$, based on a softmax over distances to the prototypes:
\begin{equation}
		p(y=m|\boldsymbol{r}_{j}) = \frac{\exp\big(\zeta(\boldsymbol{r}_{j},\boldsymbol{\mu}_{m})\big)}{\sum_{c=1}^C\exp\big(\zeta(\boldsymbol{r}_{j},\boldsymbol{\mu}_c)\big)}.
  \label{e10}
\end{equation}
The \textit{global dispersion} loss, $\mathcal{L}_g$, thus can be expressed as
\begin{equation}
        {\mathcal{L}_g} =  - \frac{1}{N_s\times k} \sum_{{i}=1}^{N_s\times k} \lambda_{i} \log\big(p(y=y_{i}|\boldsymbol{r}_{i})\big),
        \label{ee}
\end{equation}
where $\lambda_{i}$ is used to constrain the contribution of region $i$.
We experimentally found that using different collections of region embeddings, rather than a fixed set of image embeddings (i.e. $\{\boldsymbol{e}_{i}\}_{{}=1}^{N_s}$) as queries to enlarge distances among image-level class prototypies in different iterations is more effective (in Eq. \ref{e10}). One possible reason is that in addition to promoting the inter-class dispersion of {clean}, image-level class prototypes, $\mathcal{L}_g$ also complements $\mathcal{L}_l$ to improve the intra-class compactness of clean regions by Eq. \ref{e10}.

Finally, the two losses are complementary to strengthen the mining of more discriminative representations from support samples, by optimizing the following objective:
	\begin{equation} 
	\begin{aligned}
	        \mathcal{L} &= \beta \mathcal{L}_{l} + \mathcal{L}_{g}, \\
	\end{aligned}
	\label{eq12}
	\end{equation}
where $\beta$ is used to balance the importance of the two losses. 

\noindent\subsection{Task Adaptation and Inference}
During task adaptation, we iteratively construct a set of local regions from the inner-task support samples, and perform SGD update using $\mathcal{L}$.  
At inference, we only retain the adapted model to produce \textit{image representations} of support samples, on which we build a noise-robust prototypical classifier guided by the refined image weights in the momentum accumulator. More details are in \textbf{{Sup. Mat. (A)}}.

\keypoint{Discussion.}
In terms of computational efficiency, DETA is better equipped to handle the dual noises in few-shot tasks than in {training-time} base classes or other generic scenarios with large training datasets.
Computational issues in DETA caused by 
\textbf{1)} the weighting of inner-task images and regions, and 
\textbf{2)} the multiplicative expansion of support samples (brought by cropped regions) for task adaptation, 
can be substantially weakened due to the much smaller number of samples in few-shot tasks.
Please refer to \textbf{{Sup. Mat. (D)}} for an analysis of DETA w.r.t. computational efficiency.

\section{Experiments}
In this section, we perform extensive experiments to demonstrate the flexibility and effectiveness of DETA. 

\keypoint{Datasets}. 
We conduct experiments on Meta-Dataset (MD) \cite{triantafillou2019meta}, the most comprehensive and challenging large-scale FSL benchmark, which subsumes ten image datasets from various vision domains in one collection, including ImN-MD, Omglot, \textit{etc}. 
Please refer to \cite{triantafillou2019meta} for details of MD. 
We consider two versions of MD in our experiments.
\textit{Vanilla MD for image-denoising}: 
We assume the labels of the ten vanilla MD datasets are clean -- a commonly-used assumption in generic label-denoising tasks \cite{liang2022few,kong2021resolving,frenay2013classification}, and directly use the vanilla MD to verify the image-denoising performance of our method.
\textit{Label-corrupted MD for label-denoising}: 
following \cite{liang2022few,kong2021resolving} we scrutinize the robustness of DETA to label noise, by manually corrupting the labels of various ratios (10\%$\sim$70\%) of support samples.
Yet, it is worth mentioning that in the standard task-sampling protocol for MD, the generated test tasks are way/shot imbalanced, \textit{a.k.a.} \textit{varied-way varied-shot}. 
To avoid cases where the number of support samples in a class is less than 10, we adopt a unified task sampling protocol for the two MD versions by fixing the shot of every inner-task class to 10, \textit{i.e.}, \textit{varied-way 10-shot}. 
However, when conducting comparisons with state-of-the-arts, we still employ the standard {varied-way varied-shot} protocol for fair comparison.

\keypoint{Baseline Methods.}
We verify the effectiveness and flexibility of DETA by applying it to a broad spectrum of baseline methods applied on various  diverse pre-trained models.  
For $\mathtt{A}$-$\mathtt{TA}$, we consider the two strong baselines {TSA} \cite{li2022cross} and {eTT} \cite{xuexploring}.
Both of them integrate a model-specific adapter to the pre-trained model: 
TSA integrates a residual adapter to the single-domain URL (w/ RN-18 pre-trained on $84\times84$ ImN-MD) \cite{li2021universal} and eTT attaches a self-attention adapter to DINO (ViT-S) \cite{caron2021emerging}.
As for $\mathtt{F}$-$\mathtt{TA}$, motivated by \cite{li2022cross}\footnote{The nonparametric NCC has been proven in \cite{li2022cross} to be more effective for adapter-based or finetuning-based task adaptation than other competitors such as logistic regression, support vector machine and Mahal. Dist.}, 
we use the NCC head instead of a linear classifier which is common in transfer learning literature. 
We denote this $\mathtt{F}$-$\mathtt{TA}$ scheme F-NCC, and use it for adapting different pre-trained models including MOCO (w/ RN-50) \cite{he2020momentum}, CLIP (w/ RN-50) \cite{radford2021learning}, DeiT (w/ ViT-S) \cite{deit} and Swin Transformer (Tiny) \cite{liu2021swin}. 
All models are trained on Imagenet-1k, except for CLIP, which is trained on large-scale image captions.
For all baseline methods, we match the image size in model pre-training and task adaptation, \textit{i.e.}, the image size  is set to $84\times84$ for \textit{TSA} \cite{li2022cross}, and $224\times224$ for other methods.

\keypoint{Implementation Details.}
Following \cite{xuexploring,li2022cross}, we perform task adaptation by updating the pre-trained model (or the appended task adapter) for 40 iterations on each few-shot task.
During each iteration of our DETA, 4 and 2 image regions are cropped from every support sample for TSA and other methods, respectively. 
The projection head in our network is a two-layer MLP, and the embedding dimension is 128.
The two temperatures $\tau$ and $\pi$, are set to 0.5 and 0.07, respectively. 
The hyperparameter $\beta$ is set to 0.1.
More detailed settings are provided in \textbf{Sup. Mat. (B)}.

\keypoint{Evaluation Metric.}
We evaluate our method on 600 randomly sampled test tasks for each MD dataset, and report average accuracy (in \%) and 95\% confidence intervals.

\subsection{Experimental Results}
In this part, we seek to answer the following questions.

\keypoint{Q1.} Can DETA consistently enhance task adaptation results for any types of baselines by performing image-denoising on support samples?

\keypoint{Q2.} Can DETA perform robustly in the presence of various ratios of label-noisy support samples?

\keypoint{Q3.} Can DETA boost the current state-of-the-art, after tackling the overlooked image noise in the MD benchmark?

\begin{table*}[htbp]
\setlength{\abovecaptionskip}{0.1cm}  
\setlength{\belowcaptionskip}{-0.2cm} 
  \centering
  \tabcolsep 0.042in
  \footnotesize
    \begin{tabular}{c|c|cccccccccc|c}
    \Xhline{1.5 pt}
    \multicolumn{1}{c|}{{Model}} & \multicolumn{1}{c|}{Method}  &ImN-MD & Omglot & Acraft&CUB& DTD& QkDraw& Fungi& Flower& COCO&Sign& {\textit{Avg}} \\
    \hline 
    URL & \textit{TSA} \cite{li2022cross}   &58.3\scriptsize{ ± 0.9}  &80.7\scriptsize{ ± 0.2}  &61.1\scriptsize{ ± 0.7} &83.2\scriptsize{ ± 0.5}& 72.5\scriptsize{ ± 0.6} & 78.9\scriptsize{ ± 0.6} & 64.7\scriptsize{ ± 0.8} & 92.3\scriptsize{ ± 0.3}  & 75.1\scriptsize{ ± 0.7}& 87.7\scriptsize{ ± 0.4}  & 75.5  \\
     {(RN-18)} &\cellcolor{gray!15} \textbf{+ \textit{DETA}} &\cellcolor{gray!15}\textbf{58.7\scriptsize{ ± 0.9}}  &\cellcolor{gray!15}\textbf{82.7\scriptsize{ ± 0.2}} &\cellcolor{gray!15}\textbf{63.1\scriptsize{ ± 0.7}} &\cellcolor{gray!15}\textbf{85.0\scriptsize{ ± 0.5}}&\cellcolor{gray!15} \textbf{72.7\scriptsize{ ± 0.6}} &\cellcolor{gray!15} \textbf{80.4\scriptsize{ ± 0.6}} &\cellcolor{gray!15} \textbf{66.7\scriptsize{ ± 0.8}} &\cellcolor{gray!15} \textbf{93.8\scriptsize{ ± 0.3}}  &\cellcolor{gray!15} \textbf{76.3\scriptsize{ ± 0.7}}&\cellcolor{gray!15} \textbf{92.1\scriptsize{ ± 0.4}}  &\cellcolor{gray!15} \textbf{77.3} (\textcolor{blue}{\scriptsize{+1.8}})  \\
     \hline
     DINO & \textit{eTT} \cite{xuexploring}  &73.2\scriptsize{ ± 0.8}  & 93.0\scriptsize{ ± 0.4} & \textbf{68.1\scriptsize{ ± 0.7}}  & 89.6\scriptsize{ ± 0.3}& 74.9\scriptsize{ ± 0.5}  & 79.3\scriptsize{ ± 0.7} & 76.2\scriptsize{ ± 0.5}  & 96.0\scriptsize{ ± 0.2} & 72.7\scriptsize{ ± 0.6}  & 86.3\scriptsize{ ± 0.7}  & 80.9 \\
     {(ViT-S)} &\cellcolor{gray!15} \textbf{+ \textit{DETA}} &\cellcolor{gray!15}\textbf{75.6\scriptsize{ ± 0.8}} &\cellcolor{gray!15}\textbf{93.6\scriptsize{ ± 0.4}} &\cellcolor{gray!15}{67.7\scriptsize{ ± 0.8}} &\cellcolor{gray!15}\textbf{91.8\scriptsize{ ± 0.3}} &\cellcolor{gray!15}\textbf{76.0\scriptsize{ ± 0.5}} &\cellcolor{gray!15}\textbf{81.9\scriptsize{ ± 0.7}} &\cellcolor{gray!15}\textbf{77.2\scriptsize{ ± 0.5}} &\cellcolor{gray!15}\textbf{96.9\scriptsize{ ± 0.3}} &\cellcolor{gray!15}\textbf{78.5\scriptsize{ ± 0.6}} &\cellcolor{gray!15}\textbf{88.5\scriptsize{ ± 0.7}} &\cellcolor{gray!15} \textcolor{black}{\textbf{82.8}} (\textcolor{blue}{\scriptsize{+1.9}}) \\
    \hline
    MoCo  & \textit{F-NCC} & 70.7\scriptsize{ ± 1.0} &82.5\scriptsize{ ± 0.4} & 55.1\scriptsize{ ± 0.8} & 67.0\scriptsize{ ± 0.8} &  \textcolor{black}{\textbf{81.3\scriptsize{ ± 0.5}}} & 73.8\scriptsize{ ± 0.7} & 54.8\scriptsize{ ± 0.9} & 89.2\scriptsize{ ± 0.5}& 76.8\scriptsize{ ± 0.7}  & 79.6\scriptsize{ ± 0.6}  & 73.0  \\
    (RN-50) &\cellcolor{gray!15} \textbf{+ \textit{DETA}}  & \cellcolor{gray!15} \textcolor{black}{\textbf{73.6\scriptsize{ ± 1.0}}}  & \cellcolor{gray!15}\textcolor{black}{\textbf{83.9\scriptsize{ ± 0.4}}}  & \cellcolor{gray!15} \textcolor{black}{\textbf{59.1\scriptsize{ ± 0.8}}}  & \cellcolor{gray!10} \textcolor{black}{\textbf{73.9\scriptsize{ ± 0.8}}}  & \cellcolor{gray!15}80.9\scriptsize{ ± 0.5} & \cellcolor{gray!15} \textcolor{black}{\textbf{76.1\scriptsize{ ± 0.7}}}  &\cellcolor{gray!15}  \textcolor{black}{\textbf{60.7\scriptsize{ ± 0.9}}} & \cellcolor{gray!15} \textcolor{black}{\textbf{92.3\scriptsize{ ± 0.5}}} & \cellcolor{gray!15}\textcolor{black}{\textbf{79.0\scriptsize{ ± 0.7}}}    & \cellcolor{gray!15} \textcolor{black}{\textbf{84.2\scriptsize{ ± 0.6}}}   & \cellcolor{gray!15} \textcolor{black}{\textbf{76.4}}  (\textcolor{blue}{\scriptsize{+3.4}})  \\
    \hline  
    CLIP  &  \textit{F-NCC} & 67.0\scriptsize{ ± 1.0} &89.2\scriptsize{ ± 0.5} & 61.2\scriptsize{ ± 0.8} & 84.0\scriptsize{ ± 0.7} & 74.5\scriptsize{ ± 0.6} & 75.5\scriptsize{ ± 0.7} & 57.6\scriptsize{ ± 0.9} & 92.1\scriptsize{ ± 0.4}  & 72.1\scriptsize{ ± 0.8}& 79.8\scriptsize{ ± 0.7}  & 75.3 \\
    (RN-50) &\cellcolor{gray!15}  \textbf{+ \textit{DETA}} & \cellcolor{gray!15}\textcolor{black}{\textbf{69.6\scriptsize{ ± 0.9}}} &\cellcolor{gray!15}\textcolor{black}{\textbf{92.2\scriptsize{ ± 0.5}}} &\cellcolor{gray!15} \textcolor{black}{\textbf{59.7\scriptsize{ ± 0.8}}} &\cellcolor{gray!15} \textcolor{black}{\textbf{88.5\scriptsize{ ± 0.7}}} &\cellcolor{gray!15}\textcolor{black}{\textbf{76.2\scriptsize{ ± 0.6}}} &\cellcolor{gray!15}\textcolor{black}{\textbf{77.2\scriptsize{ ± 0.7}}}  &\cellcolor{gray!15}\textcolor{black}{\textbf{64.5\scriptsize{ ± 0.9}}}  &\cellcolor{gray!15} \textcolor{black}{\textbf{94.5\scriptsize{ ± 0.3}}} &\cellcolor{gray!15} \textcolor{black}{\textbf{72.6\scriptsize{ ± 0.8}}}  &\cellcolor{gray!15}\textcolor{black}{\textbf{80.7\scriptsize{ ± 0.7}}}   &\cellcolor{gray!15}\textcolor{black}{\textbf{77.6}}  (\textcolor{blue}{\scriptsize{+2.3}}) \\ 
    \hline
    DeiT& \textit{F-NCC} & 90.0\scriptsize{ ± 0.6}&92.5\scriptsize{ ± 0.2}&65.3\scriptsize{ ± 0.7}&89.8\scriptsize{ ± 0.4}&73.9\scriptsize{ ± 0.6}&83.3\scriptsize{ ± 0.5}&70.3\scriptsize{ ± 0.8}&92.2\scriptsize{ ± 0.4}&83.0\scriptsize{ ± 0.6}&85.0\scriptsize{ ± 0.6}&82.5  \\ 
    (ViT-S) &\cellcolor{gray!15} \textbf{+ \textit{DETA}} &\cellcolor{gray!15} \textcolor{black}{\textbf{90.8\scriptsize{ ± 0.6}}} &\cellcolor{gray!15}\textcolor{black}{\textbf{93.3\scriptsize{ ± 0.2}}} & \cellcolor{gray!15}\textcolor{black}{\textbf{71.6\scriptsize{ ± 0.7}}} &\cellcolor{gray!15} \textcolor{black}{\textbf{92.4\scriptsize{ ± 0.4}}} &\cellcolor{gray!15}\textcolor{black}{\textbf{78.0\scriptsize{ ± 0.6}}} &\cellcolor{gray!15}\textcolor{black}{\textbf{84.1\scriptsize{ ± 0.6}}}  &\cellcolor{gray!15}\textcolor{black}{\textbf{75.2\scriptsize{ ± 0.8}}} &\cellcolor{gray!15} \textcolor{black}{\textbf{84.4\scriptsize{ ± 0.4}}}   &\cellcolor{gray!15} \textcolor{black}{\textbf{95.5\scriptsize{ ± 0.6}}} &\cellcolor{gray!15}\textcolor{black}{\textbf{90.0\scriptsize{ ± 0.6}}}  &\cellcolor{gray!15}\textcolor{black}{\textbf{85.2}}   (\textcolor{blue}{\scriptsize{+2.7}}) \\
    \hline
    Vanilla & \textit{F-NCC} & 90.8\scriptsize{ ± 0.8}&91.2\scriptsize{ ± 0.3}&57.6\scriptsize{ ± 1.0}&88.3\scriptsize{ ± 0.5}&76.4\scriptsize{ ± 0.6}&81.9\scriptsize{ ± 0.8}&67.8\scriptsize{ ± 0.9}&92.3\scriptsize{ ± 0.4}&82.5\scriptsize{ ± 0.6}&83.9\scriptsize{ ± 0.8}&81.3  \\ 
    SwinT&\cellcolor{gray!15} \textbf{+ \textit{DETA}} &\cellcolor{gray!15} \textcolor{black}{\textbf{91.8\scriptsize{ ± 0.9}}} &\cellcolor{gray!15}\textcolor{black}{\textbf{92.5\scriptsize{ ± 0.3}}} &\cellcolor{gray!15} \textcolor{black}{\textbf{68.9\scriptsize{ ± 0.9}}} &\cellcolor{gray!15} \textcolor{black}{\textbf{92.7\scriptsize{ ± 0.5}}} &\cellcolor{gray!15}\textcolor{black}{\textbf{79.5\scriptsize{ ± 0.7}}} &\cellcolor{gray!15}\textcolor{black}{\textbf{82.8\scriptsize{ ± 0.6}}}  &\cellcolor{gray!15}\textcolor{black}{\textbf{76.6\scriptsize{ ± 0.8}}}  &\cellcolor{gray!15} \textcolor{black}{\textbf{96.4\scriptsize{ ± 0.4}}} &\cellcolor{gray!15} \textcolor{black}{\textbf{82.9 \scriptsize{± 0.4}}}   &\cellcolor{gray!15}\textcolor{black}{\textbf{89.9\scriptsize{ ± 0.7}}}  &\cellcolor{gray!15}\textcolor{black}{\textbf{85.4}}   (\textcolor{blue}{\scriptsize{+4.1}}) \\
    \hline 
    \end{tabular}%
    \caption{Few-shot classification results of different methods on MD.
    The $\mathtt{A}$-$\mathtt{TA}$ methods \textit{TSA}\cite{li2022cross} and {\textit{eTT}} \cite{xuexploring} integrate a model-specific adapter to the pre-trained model, while the $\mathtt{F}$-$\mathtt{TA}$ method \textit{F-NCC} use a model-agnostic NCC head for adapting different pre-trained models.
    }
  \label{t1}%
\end{table*}%
\begin{table}[htbp]
\setlength{\abovecaptionskip}{0.1cm}  
  \centering 
  \tabcolsep 0.04in
  \footnotesize
    \begin{tabular}{c|c|cccc}
    \Xhline{1.5 pt} 
    \multicolumn{1}{c|}{\multirow{2}{*}{\makecell[c]{{Model}}}} & \multicolumn{1}{c|}{\multirow{2}{*}{\makecell[c]{{Method}}}}  & \multicolumn{4}{c}{\textit{Ratio of noisy labels}}    \\
    \cline{3-6} 
    &&  \multicolumn{1}{c}{10\%} & \multicolumn{1}{c}{30\%} & \multicolumn{1}{c}{50\%} & \multicolumn{1}{c}{70\%}  \\
    \hline
    \multicolumn{1}{c|}{\multirow{2}{*}{\makecell[c]{{URL}\\
    \,(RN-18)\,}}}& \textit{TSA} \cite{li2022cross} &72.8&65.0&54.1&38.3 \\
     &\cellcolor{gray!15}  \textbf{+ \textit{DETA}}  &\cellcolor{gray!15}\textbf{74.8} (\textcolor{blue}{\scriptsize{+2.0}})&\cellcolor{gray!15}\textbf{67.2} (\textcolor{blue}{\scriptsize{+2.2}})&\cellcolor{gray!15}\textbf{56.0} (\textcolor{blue}{\scriptsize{+1.9}}) &\cellcolor{gray!15}\textbf{40.1} (\textcolor{blue}{\scriptsize{+1.8}}) \\
    \hline
    \multicolumn{1}{c|}{\multirow{2}{*}{\makecell[c]{{DINO}\\
    \,(ViT-S)\,}}}& \textit{eTT} \cite{xuexploring} &78.0&67.7&53.8&37.8 \\
     &\cellcolor{gray!15}  \textbf{+ \textit{DETA}}  &
     \cellcolor{gray!15}\textbf{80.3} (\textcolor{blue}{\scriptsize{+2.3}})&\cellcolor{gray!15}\textbf{70.7} (\textcolor{blue}{\scriptsize{+3.0}})&\cellcolor{gray!15}\textbf{58.0} (\textcolor{blue}{\scriptsize{+4.2}})&\cellcolor{gray!15}\textbf{41.9} (\textcolor{blue}{\scriptsize{+4.1}}) \\
    \hline
    \multicolumn{1}{c|}{\multirow{2}{*}{\makecell[c]{{MoCo}\\
    (RN-50)}}} &  \textit{F-NCC} &70.4&63.3&52.4&36.6 \\
     &\cellcolor{gray!15}  \textbf{+ \textit{DETA}} &\cellcolor{gray!15}\textbf{74.1} (\textcolor{blue}{\scriptsize{+3.7}})&\cellcolor{gray!15}\textbf{68.0} (\textcolor{blue}{\scriptsize{+4.7}})&\cellcolor{gray!15}\textbf{57.8} (\textcolor{blue}{\scriptsize{+5.4}}) &\cellcolor{gray!15}\textbf{40.1} (\textcolor{blue}{\scriptsize{+3.5}}) \\
    \hline 
    \multicolumn{1}{c|}{\multirow{2}{*}{\makecell[c]{{CLIP}\\
    (RN-50)}}}  & \textit{F-NCC} &73.0&65.5&53.3&36.9  \\
     & \cellcolor{gray!15}  \textbf{+ \textit{DETA}} &\cellcolor{gray!15}\textbf{75.7} (\textcolor{blue}{\scriptsize{+2.7}})& \cellcolor{gray!15}\textbf{69.7} (\textcolor{blue}{\scriptsize{+4.2}}) &\cellcolor{gray!15}\textbf{58.5} (\textcolor{blue}{\scriptsize{+5.2}}) &\cellcolor{gray!15}\textbf{40.8} (\textcolor{blue}{\scriptsize{+3.9}})\\
    \hline
    \multicolumn{1}{c|}{\multirow{2}{*}{\makecell[c]{DeiT \\
    (ViT-S)}}} &  \textit{F-NCC}  &80.0&74.3&64.1&44.9 \\
     & \cellcolor{gray!15} \textbf{+ \textit{DETA}}  &\cellcolor{gray!15}\textbf{83.3} (\textcolor{blue}{\scriptsize{+3.3}})&\cellcolor{gray!15}\textbf{77.2} (\textcolor{blue}{\scriptsize{+2.9}})&\cellcolor{gray!15} \textbf{67.1} (\textcolor{blue}{\scriptsize{+3.0}})&\cellcolor{gray!15}\textbf{47.7} (\textcolor{blue}{\scriptsize{+2.8}}) \\
    \hline
    \multicolumn{1}{c|}{\multirow{2}{*}{\makecell[c]{Vanilla\\
    SwinT}}} & \textit{F-NCC}  &78.8&71.6&59.8&42.2  \\
     & \cellcolor{gray!15} \textbf{+ \textit{DETA}}  &\cellcolor{gray!15}\textbf{83.9} (\textcolor{blue}{\scriptsize{+5.1}})&\cellcolor{gray!15}\textbf{77.3} (\textcolor{blue}{\scriptsize{+5.7}})&\cellcolor{gray!15} \textbf{65.9} (\textcolor{blue}{\scriptsize{+6.1}})&\cellcolor{gray!15}\textbf{46.8} (\textcolor{blue}{\scriptsize{+4.6}})\\
    \hline 
    \end{tabular}%
    \caption{Average few-shot classification results of different models on MD, with various ratios of label-noisy support samples. 
    }
  \label{table2}%
\end{table}%

\begin{table}[tp]
\setlength{\abovecaptionskip}{0.1cm}  
		\centering
		 \tabcolsep 0.115in
		\footnotesize
		\begin{tabular}{cc|cc}
			\Xhline{1.5 pt}
			 Image & Region & Image-denoising & Label-denoising (30\%) \\
			\hline
			\scriptsize{\Checkmark} &\scriptsize{\XSolidBrush} &73.0 &63.3\\
			\rowcolor{gray!15} \scriptsize{\Checkmark} &\scriptsize{\Checkmark}& 73.8 (\textcolor{blue}{\scriptsize{+0.8}})  &63.9 (\textcolor{blue}{\scriptsize{+0.6}}) \\
			\hline
		\end{tabular}
		\caption{The impact of data augmentation caused by cropped regions on model performance. The baseline is MoCo (w/ RN-50).}
		\label{ag}
\end{table}

\begin{table}[tp]
\setlength{\abovecaptionskip}{0.1cm}  
\setlength{\belowcaptionskip}{-0.2cm} 
		\centering
		 \tabcolsep 0.0551in
		\footnotesize
  		\begin{tabular}{c|l|cc}
			\Xhline{1.5 pt}
			 ID & \quad \quad \quad \quad \quad \quad  Setting & Img-denois. & Label-denois. \\
			 \hline
            \scriptsize$\mathbb{A}$ & Baseline &73.8 &63.9 \\
            \hline
            \scriptsize$\mathbb{B}$ & + CoRA$^*$ &74.1 &64.3 \\
            \scriptsize$\mathbb{C}$ & + CoRA &74.4 &64.8 \\
            \scriptsize$\mathbb{D}$ & +  CoRA + $\mathcal{L}_l$ &75.5 &65.4 \\
            \scriptsize$\mathbb{E}$ & +  CoRA + $\mathcal{L}_g$ &75.2 &66.6 \\
            \scriptsize$\mathbb{F}$ & +  CoRA + $\mathcal{L}_l$ + $\mathcal{L}_g$ &75.8 &67.1 \\
            \rowcolor{gray!15}
             \scriptsize $\mathbb{G}$ & +  CoRA + $\mathcal{L}_l$ + $\mathcal{L}_g$ + MA ({{DETA}}) &\textbf{76.4} (\textcolor{blue}{\scriptsize{+2.6}}) &\textbf{68.0} (\textcolor{blue}{\scriptsize{+4.1}})\\
			\hline
		\end{tabular}
		\caption{Ablation studies for the designed components of DETA on MD. The baseline model is MoCo (w/ RN-50), the ratio of label-noisy support samples is set to 30\%.}
		\label{ab}
\end{table}

\begin{table*}[htbp]
\setlength{\abovecaptionskip}{0.1cm}  
  \centering
  \tabcolsep 0.054in
  \footnotesize
  \begin{threeparttable}
    \begin{tabular}{l|cccccccccc|c}
    \Xhline{1.5 pt}
    \multicolumn{1}{l|}{\multirow{2}{*}{\makecell[l]{Method (w/ RN-18) $ ^{\clubsuit} $}}} & \multicolumn{1}{c}{\textit{In-Domain}}   & \multicolumn{9}{|c|}{\textit{Out-of-Domain}} & \multicolumn{1}{c}{\multirow{2}{*}{\textit{Avg}}} \\ 
    \cline{2-11} 
    \multicolumn{1}{c|}{} & \multicolumn{1}{c}{ImN-MD} & \multicolumn{1}{|c}{Omglot} & \multicolumn{1}{c}{Acraft} & \multicolumn{1}{c}{CUB} & \multicolumn{1}{c}{DTD} & \multicolumn{1}{c}{QkDraw} & \multicolumn{1}{c}{Fungi} & \multicolumn{1}{c}{Flower} & \multicolumn{1}{c}{COCO} & \multicolumn{1}{c|}{Sign} &  \\
    \hline
    Finetune \cite{triantafillou2019meta} &45.8\scriptsize{ ± 1.1}&60.9\scriptsize{ ± 1.6}&68.7 ± 1.3&57.3\scriptsize{ ± 1.3}&69.0\scriptsize{ ± 0.9}&42.6\scriptsize{ ± 1.2}&38.2\scriptsize{ ± 1.0}&85.5\scriptsize{ ± 0.7}&34.9\scriptsize{ ± 1.0}&66.8\scriptsize{ ± 1.3}&57.0  \\
    ProtoNet \cite{triantafillou2019meta} &50.5\scriptsize{ ± 1.1}&60.0\scriptsize{ ± 1.4}&53.1\scriptsize{ ± 1.0}&68.8\scriptsize{ ± 1.0}&66.6\scriptsize{ ± 0.8}&49.0\scriptsize{ ± 1.1}&39.7\scriptsize{ ± 1.1} &85.3\scriptsize{ ± 0.8}&41.0\scriptsize{ ± 1.1}&47.1\scriptsize{ ± 1.1}&56.1  \\
    FoProMA \cite{triantafillou2019meta} &49.5\scriptsize{ ± 1.1}&63.4\scriptsize{ ± 1.3}&56.0\scriptsize{ ± 1.0}&68.7\scriptsize{ ± 1.0}&66.5\scriptsize{ ± 0.8}&51.5\scriptsize{ ± 1.0}&40.0\scriptsize{ ± 1.1}&87.2\scriptsize{ ± 0.7}&43.7\scriptsize{ ± 1.1}&48.8\scriptsize{ ± 1.1}&57.5  \\
    Alfa-FoProMA \cite{triantafillou2019meta} &52.8\scriptsize{ ± 1.1}&61.9\scriptsize{ ± 1.5}&63.4\scriptsize{ ± 1.1}&69.8\scriptsize{ ± 1.1}&70.8\scriptsize{ ± 0.9}&59.2\scriptsize{ ± 1.2}&41.5\scriptsize{ ± 1.2}&86.0\scriptsize{ ± 0.8}&48.1\scriptsize{ ± 1.1}&60.8\scriptsize{ ± 1.3}&61.4 \\
    BOHB \cite{saikia2020optimized}  &51.9\scriptsize{ ± 1.1}&67.6\scriptsize{ ± 1.2}&54.1\scriptsize{ ± 0.9}&70.7\scriptsize{ ± 0.9}&68.3\scriptsize{ ± 0.8}&50.3\scriptsize{ ± 1.0}&41.4\scriptsize{ ± 1.1}&87.3\scriptsize{ ± 0.6}&48.0\scriptsize{ ± 1.0}&51.8\scriptsize{ ± 1.0}&59.1  \\
    FLUTE \cite{triantafillou2021learning} &46.9\scriptsize{ ± 1.1}&61.6\scriptsize{ ± 1.4}&48.5\scriptsize{ ± 1.0}&47.9\scriptsize{ ± 1.0}&63.8\scriptsize{ ± 0.8 }&57.5\scriptsize{ ± 1.0}&31.8\scriptsize{ ± 1.0}&80.1\scriptsize{ ± 0.9}&41.4\scriptsize{ ± 1.0}&46.5\scriptsize{ ± 1.1}&52.6  \\
    eTT$^{\sharp}$ \cite{xuexploring} &56.4\scriptsize{ ± 1.1}&72.5\scriptsize{ ± 1.4}&72.8\scriptsize{ ± 1.0}&73.8\scriptsize{ ± 1.1}&77.6\scriptsize{ ± 0.8}&68.0\scriptsize{ ± 0.9}&\textbf{51.2\scriptsize{ ± 1.1}}&\textbf{93.3\scriptsize{ ± 0.6}}&55.7\scriptsize{ ± 1.0}&84.1\scriptsize{ ± 1.0}&70.5  \\
    \hline
    \rowcolor{gray!15} \textit{URL} (\textit{Base Model}) \cite{li2021universal}  &56.8\scriptsize{ ± 1.0}&79.5\scriptsize{ ± 0.8}&49.4\scriptsize{ ± 0.8}&71.8\scriptsize{ ± 0.9}&72.7\scriptsize{ ± 0.7}&53.4\scriptsize{ ± 1.0}&40.9\scriptsize{ ± 0.9}&85.3\scriptsize{ ± 0.7}&52.6\scriptsize{ ± 0.9}&47.3\scriptsize{ ± 1.0}&61.1  \\
    \rowcolor{gray!15} \textit{+ Beta}\cite{li2021universal}  &58.4\scriptsize{ ± 1.1}&81.1\scriptsize{ ± 0.8}&51.9\scriptsize{ ± 0.9}&73.6\scriptsize{ ± 1.0}&74.0\scriptsize{ ± 0.7}&55.6\scriptsize{ ± 1.0}&42.2\scriptsize{ ± 0.9}&86.2\scriptsize{ ± 0.8}&55.1\scriptsize{ ± 1.0}&59.0\scriptsize{ ± 1.1}&63.7  \\
    \rowcolor{gray!15} \textit{+ TSA} \cite{li2022cross}  &59.5\scriptsize{ ± 1.1}&78.2\scriptsize{ ± 1.2}&72.2\scriptsize{ ± 1.0}&74.9\scriptsize{ ± 0.9}&77.3\scriptsize{ ± 0.7}&67.6\scriptsize{ ± 0.9 }&44.7\scriptsize{ ± 1.0}&90.9\scriptsize{ ± 0.6} &59.0\scriptsize{ ± 1.0}&82.5\scriptsize{ ± 0.8}&70.7  \\
   \rowcolor{gray!15} \textbf{\textit{+ TSA + DETA}} &\textcolor{black}{\textbf{60.7\scriptsize{ ± 1.0}}}&\textcolor{black}{\textbf{81.6\scriptsize{ ± 1.2}}}&\textcolor{black}{\textbf{73.0\scriptsize{ ± 1.0}}}&\textcolor{black}{\textbf{77.0\scriptsize{ ± 0.9}}}&\textcolor{black}{\textbf{78.3\scriptsize{ ± 0.7}}}&\textcolor{black}{\textbf{69.5\scriptsize{ ± 0.9}}}&\textcolor{black}{{47.6\scriptsize{ ± 1.0}}}&\textcolor{black}{{92.6\scriptsize{ ± 0.6}}}&\textcolor{black}{\textbf{60.3\scriptsize{ ± 1.0}}}&\textcolor{black}{\textbf{86.8\scriptsize{ ± 0.8}}}&\textcolor{black}{\textbf{72.8}}  \\
    \hline
    \multicolumn{1}{l|}{\multirow{2}{*}{\makecell[c]{Method (w/ RN-18) $ ^{\spadesuit} $}}} & \multicolumn{8}{c}{\textit{In-Domain}}   & \multicolumn{2}{|c|}{\textit{Out-of-Domain}} & \multicolumn{1}{c}{\multirow{2}{*}{\textit{Avg}}} \\
    \cline{2-11}
    \multicolumn{1}{c|}{} & \multicolumn{1}{c}{ImN-MD} & \multicolumn{1}{c}{Omglot} & \multicolumn{1}{c}{Acraft} & \multicolumn{1}{c}{CUB} & \multicolumn{1}{c}{DTD} & \multicolumn{1}{c}{QkDraw} & \multicolumn{1}{c}{Fungi} & \multicolumn{1}{c}{Flower} & \multicolumn{1}{|c}{COCO} & \multicolumn{1}{c|}{Sign} &  \\
    \hline
    CNAPS \cite{requeima2019fast}  &50.8\scriptsize{ ± 1.1}&91.7\scriptsize{ ± 0.5}&83.7\scriptsize{ ± 0.6}&73.6\scriptsize{ ± 0.9}&59.5\scriptsize{ ± 0.7}&74.7\scriptsize{ ± 0.8}&50.2\scriptsize{ ± 1.1}&88.9\scriptsize{ ± 0.5}&39.4\scriptsize{ ± 1.0}&56.5\scriptsize{ ± 1.1}&66.9  \\
    SimpCNAPS \cite{bateni2020improved} &58.4\scriptsize{ ± 1.1}&91.6\scriptsize{ ± 0.6}&82.0\scriptsize{ ± 0.7}&74.8\scriptsize{ ± 0.9}&68.8\scriptsize{ ± 0.9}&76.5\scriptsize{ ± 0.8}&46.6\scriptsize{ ± 1.0}&90.5\scriptsize{ ± 0.5}&48.9\scriptsize{ ± 1.1}&57.2\scriptsize{ ± 1.0}&69.5  \\
    TransCNAPS \cite{bateni2022enhancing} &57.9\scriptsize{ ± 1.1}&94.3\scriptsize{ ± 0.4}&84.7\scriptsize{ ± 0.5}&78.8\scriptsize{ ± 0.7}&66.2\scriptsize{ ± 0.8}&77.9\scriptsize{ ± 0.6}&48.9\scriptsize{ ± 1.2}&92.3\scriptsize{ ± 0.4}&42.5\scriptsize{ ± 1.1}&59.7\scriptsize{ ± 1.1}&70.3  \\
    SUR \cite{dvornik2020selecting}   &56.2\scriptsize{ ± 1.0}&94.1\scriptsize{ ± 0.4}&85.5\scriptsize{ ± 0.5}&71.0\scriptsize{ ± 1.0}&71.0\scriptsize{ ± 0.8}&81.8\scriptsize{ ± 0.6}&64.3\scriptsize{ ± 0.9}&82.9\scriptsize{ ± 0.8}&52.0\scriptsize{ ± 1.1}&51.0\scriptsize{ ± 1.1}&71.0  \\
    URT \cite{liu2020universal}  &56.8\scriptsize{ ± 1.1}&94.2\scriptsize{ ± 0.4}&85.8\scriptsize{ ± 0.5}&76.2\scriptsize{ ± 0.8}&71.6\scriptsize{ ± 0.7}&82.4\scriptsize{ ± 0.6}&64.0\scriptsize{ ± 1.0}&87.9\scriptsize{ ± 0.6}&48.2\scriptsize{ ± 1.1}&51.5\scriptsize{ ± 1.1}&71.9  \\
    FLUTE \cite{triantafillou2021learning} &58.6\scriptsize{ ± 1.0}&92.0\scriptsize{ ± 0.6}&82.8\scriptsize{ ± 0.7}&75.3\scriptsize{ ± 0.8}&71.2\scriptsize{ ± 0.8}&77.3\scriptsize{ ± 0.7}&48.5\scriptsize{ ± 1.0}&90.5\scriptsize{ ± 0.5}&52.8\scriptsize{ ± 1.1}&63.0\scriptsize{ ± 1.0}&71.2  \\
    Tri-M \cite{liu2021multi} &51.8\scriptsize{ ± 1.1}&93.2\scriptsize{ ± 0.5}&87.2\scriptsize{ ± 0.5}&79.2\scriptsize{ ± 0.8}&68.8\scriptsize{ ± 0.8}&79.5\scriptsize{ ± 0.7}&58.1\scriptsize{ ± 1.1}&91.6\scriptsize{ ± 0.6}&50.0\scriptsize{ ± 1.0}&58.4\scriptsize{ ± 1.1}&71.8  \\
    \hline
    \rowcolor{gray!15} \textit{URL} (\textit{Base Model}) \cite{li2021universal}  &57.0\scriptsize{ ± 1.0}&94.4\scriptsize{ ± 0.4}&88.0\scriptsize{ ± 0.5}&80.3\scriptsize{ ± 0.7}&74.6\scriptsize{ ± 0.7}&81.8\scriptsize{ ± 0.6}&66.2\scriptsize{ ± 0.9}&91.5\scriptsize{ ± 0.5}&54.1\scriptsize{ ± 1.0}&49.8\scriptsize{ ± 1.0}&73.8 \\
    \rowcolor{gray!15} \textit{+ Beta} \cite{li2021universal}  &58.8\scriptsize{ ± 1.1}&94.5\scriptsize{ ± 0.4}&89.4\scriptsize{ ± 0.4}&80.7\scriptsize{ ± 0.8}&77.2\scriptsize{ ± 0.7}&82.5\scriptsize{ ± 0.6}&68.1\scriptsize{ ± 0.9}&92.0\scriptsize{ ± 0.5}&57.3\scriptsize{ ± 1.0}&63.3\scriptsize{ ± 1.1}&76.4  \\
    \rowcolor{gray!15} \textit{+ TSA } \cite{li2022cross}   &59.5\scriptsize{ ± 1.0}&94.9\scriptsize{ ± 0.4}&89.9\scriptsize{ ± 0.4}&81.1\scriptsize{ ± 0.8}&77.5\scriptsize{ ± 0.7}&81.7\scriptsize{ ± 0.6}&66.3\scriptsize{ ± 0.8}&92.2\scriptsize{ ± 0.5}&57.6\scriptsize{ ± 1.0}&82.8\scriptsize{ ± 1.0}&78.3  \\
    \rowcolor{gray!15} \textbf{\textit{+ TSA + DETA}}  &\textcolor{black}{\textbf{61.0\scriptsize{ ± 1.0}}}&\textcolor{black}{\textbf{95.6\scriptsize{ ± 0.4}}}&\textcolor{black}{\textbf{91.4\scriptsize{ ± 0.4}}}&\textcolor{black}{\textbf{82.7\scriptsize{ ± 0.7}}}&\textcolor{black}{\textbf{78.9\scriptsize{ ± 0.7}}}&\textcolor{black}{\textbf{83.4\scriptsize{ ± 0.6}}}&\textcolor{black}{\textbf{68.2\scriptsize{ ± 0.8}}}&\textcolor{black}{\textbf{93.4\scriptsize{ ± 0.5}}}&\textcolor{black}{\textbf{58.5\scriptsize{ ± 1.0}}}&\textcolor{black}{\textbf{86.9 \scriptsize{± 1.0}}}&\textcolor{black}{\textbf{80.1}}  \\
    \hline 
    \end{tabular}
  \end{threeparttable}
  \caption{Comparison with state-of-the-arts on ten MD datasets ($84\times 84$). $ ^{\clubsuit} $ and $ ^{\spadesuit} $ indicate the 
  single-domain (trained on ImN-MD only) and multi-domain (trained on 8 datasets) settings in MD, respectively. 
  $\sharp$ means the feature backbone is ViT-T, with results copied from \cite{xuexploring}.} 
   \label{sota}%
\end{table*}%

\keypoint{Image-denoising}.
To validate the effectiveness of DETA on image-denoising, we conduct experiments on the vanilla MD with six baseline approaches shown before.
The quantitative results of the baseline methods w/ or w/o DETA are reported in Table \ref{t1}. 
We can observe from the results:
\textbf{1)} DETA consistently improves adapter-based and finetuning-based task adaptation methods, which confirms that DETA is orthogonal to those methods and able to improve model robustness to image noise for them.
\textbf{2)} DETA achieves significant performance gains on both TSA (for $84\times84$-size input images) and other methods (for $224\times224$-size images), suggesting DETA's flexibility.
\textbf{3)} DETA can tackle both the two types of image noise: \textit{background clutter} (in ImgN-MD, etc) and \textit{image corruption} (in Omglot and QkDraw), qualitative results are shown in Section \ref{qr}.

\keypoint{Label-denoising.}
We further demonstrate the effectiveness of DETA on label-denoising on the label-corrupted MD.
Concretely, we manually corrupt the labels of different ratios (10\%$\sim$70\%) of support samples for each task, by uniformly changing the correct image labels to the other $C-1$ classes.
Table \ref{table2} reports the average accuracy of different baselines methods w/ or w/o our DETA on the ten MD datasets, under different ratios of corrupted support samples.
We have the following observations.
\textbf{1)} The few-shot classification performance gradually decreases as the ratio of label-noisy support samples increases.
\textbf{2)} DETA consistently improves the baseline methods by a large margin in all settings, demonstrating its effectiveness to improve model robustness to label noise.
\textbf{3)} Compared with the obtained image-denoising results in Table \ref{t1}, the performance gains of DETA w.r.t. label-denoising are more significant.
Possible reasons are twofold. 
\textbf{i)} The negative impact of label noise on performance is more significant than that of image noise, as the label-noisy samples contain almost no valuable object features associated with the correct classes.
\textbf{ii)} When one class contains samples from other classes, our designed CoRA can identify the harmful regions more precisely by taking advantage of out-of-class relevance information.

\begin{figure*}
\setlength{\abovecaptionskip}{0.05cm}  
\setlength{\belowcaptionskip}{-0.2cm} 
		\centering
		\includegraphics[width=1\linewidth]{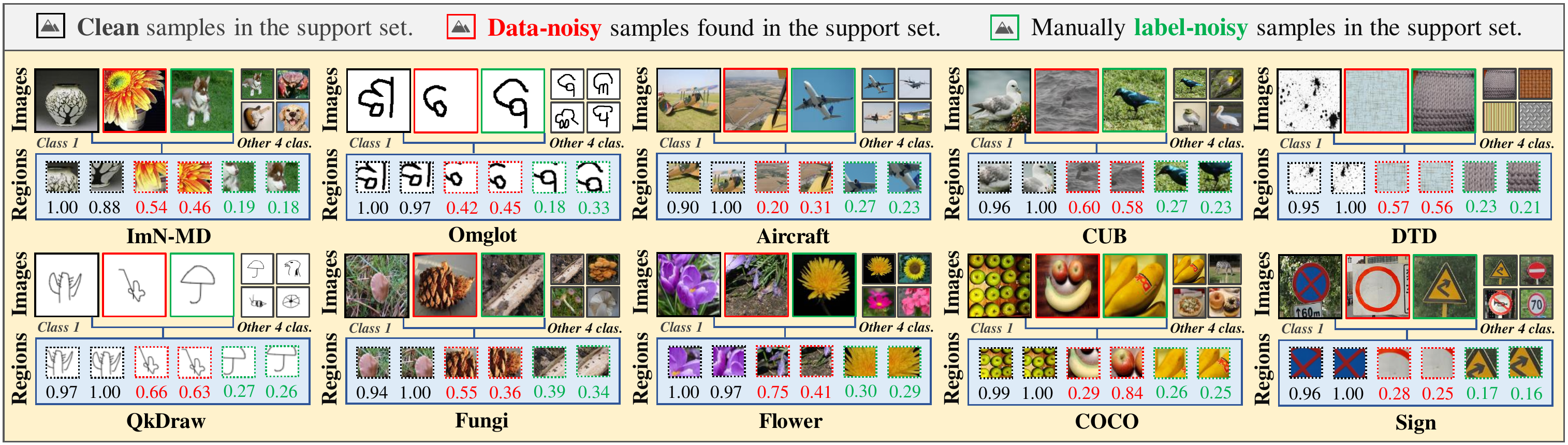} 
		\caption{Visualizations of the cropped regions and calculated weights for ten 5-way 10-shot tasks sampled from the 10 MD datasets. We record the region weights after the last iteration. To facilitate comparison, the weights in each class are divided by their maximum value.} 
		\label{vis1}
\end{figure*}

\keypoint{State-of-the-art Comparison.}
So far, we can see that our DETA can be flexibly plugged into both adapter-based and finetuning-based task adaptation methods to improve model robustness to the dual noises.
It is interesting to investigate whether DETA can further boost the current state-of-the-art after tackling the image-noisy samples in the vanilla MD.
Hence, we apply our DETA to the state-of-the-art scheme TSA\cite{li2022cross} and conduct experiments on MD with a group of competitors, \textit{e.g.}, FLUTE\cite{triantafillou2021learning}, URL\cite{li2021universal}, eTT\cite{xuexploring}. 
In Table \ref{sota}, we can observe DETA considerably improves the strong baseline TSA and establishes new state-of-the-art results on nearly all ten MD datasets, which further confirm the effectiveness and flexibility of our DETA. 
More importantly, the achieved results also uncover the ever-overlooked image noise problem of the MD benchmark. More qualitative evidence for this problem are discussed in Section \ref{qr} and demonstrated in Figure \ref{vis1}. 
 
\subsection{Ablation Studies}
\label{ablation}
Here, we conduct ablative analysis to investigate the designed components of DETA in Table \ref{ab}. 
We also study the impact of {data augmentation} caused by cropped regions on model performance in Table \ref{ag}.
Unless stated otherwise, the baseline is MoCo (w/ RN-50), the ratio of label-noisy support samples is 30\%, and the average results on the ten MD datasets are reported.

\keypoint{Influence of Data Augmentation.}
DETA leverages both the images and cropped regions of support samples to perform test-time task adaptation. 
It is important to answer the question: are the performance improvements are mostly attributed to data augmentation?
To this end, we remove all the designed components of DETA, and jointly use the images and cropped regions for task adaptation. 
The results are reported in Table \ref{ag}.
Not surprisingly, without filtering out task-irrelevant, noisy representations, the joint utilization of images and regions for task adaptation does not result in significant performance gains.

\keypoint{Effectiveness of the Designed Components.}
DETA contains three key components, 
including a CoRA module, a \textit{local compactness} loss $\mathcal{L}_l$ and a \textit{global dispersion} loss $\mathcal{L}_g$.
We conduct component-wise analysis by alternatively adding one of them to understand the influence of each component in Table \ref{ab}.
We take the random crop data augmentation as baseline ($\mathbb{A}$).
$\mathbb{B}$ or $\mathbb{C}$: only leverage CoRA to weight the support images at inference. 
CoRA$^*$: CoRA {w/o} {out-of-class relevance aggregation}.         
$\mathbb{G}$: Our DETA. “+ MA”: Inference with the {momentum accumulator}.
As seen, each component in DETA contributes to the performance.
In particular, the results in $\mathbb{F}$ suggest that $\mathcal{L}_g$ and  $\mathcal{L}_l$ complement each other to improve the denoising performance.
The results in $\mathbb{B}\&\mathbb{C}$ and $\mathbb{G}$ verify the effectiveness of the out-of-class relevance information for CoRA, and the momentum accumulator for building noise-robust classifier, respectively.

\begin{figure}
\setlength{\abovecaptionskip}{0.2cm}  
\setlength{\belowcaptionskip}{-0.3cm} 
		\centering
		\includegraphics[width=0.98\linewidth]{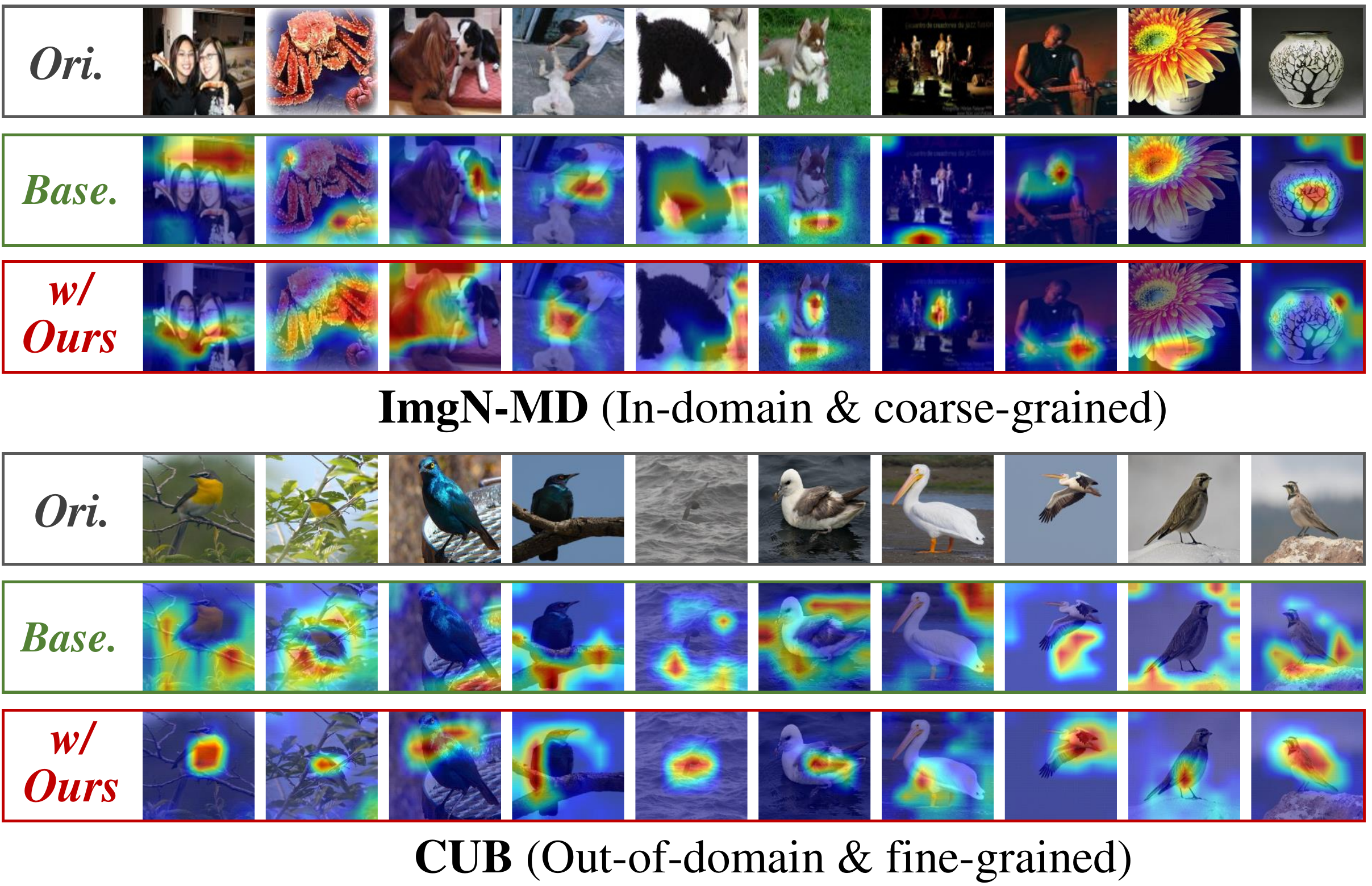} 
		\caption{CAM visualizations on two 5-way 10-shot tasks sampled from ImgN-MD and CUB, respectively. Two images per class are listed for each task. \textit{Please zoom in for details.}} 
		\label{vis2}
\end{figure}

\keypoint{Analysis of the Number of Region, Region Size, $\beta$, $\zeta(\cdot)$.}
In \textbf{Sup. Mat. (C)}, we study the impacts of the number of region, region size, $\beta$ and $\zeta(\cdot)$ on performance. We show that \textbf{1)} a too larger number of regions or a too small region size does not result in significant performance gains, and \textbf{2)} the DETA framework is in general not sensitive to $\beta$ and the choice of $\zeta(\cdot)$ within a certain range.

\noindent\subsection{Qualitative Results}
\label{qr}
Here, we provide some visualization results to qualitatively see how our method works. 
In Figure \ref{vis1}, we present the visualization of the cropped regions and the calculated weights of CoRA for few-shot tasks from MD.
As observed, CoRA successfully assigns larger (resp. smaller) weights to task-specific clean (resp. task-irrelevant noisy) regions for each task. 
In Figure \ref{vis2}, we show the CAM \cite{camvis} visualization of the activation maps for two tasks from the representative ImgN-MD and CUB. 
As shown, our method helps the baseline method accurately locate the task-specific discriminative regions in label-clean but image-noisy samples.
For example, on CUB, our method yields more attention on birds rather than cluttered backgrounds.

\section{Conclusions}
In this work, we propose DETA, a first, unified and plug-and-play framework to tackle the joint (image, label)-noise issue in test-time task adaptation. 
Without extra supervision, DETA filters out task-irrelevant, noisy representations by taking advantage of both global visual information and local region details of support sample.
We evaluate DETA on the challenging Meta-Dataset and demonstrate that it consistently improves the performance of a wide range of baseline methods applied to various pre-trained models.
We also uncover the overlooked image noise in Meta-Dataset, by tackling this issue DETA establishes new state-of-the-art results.
We hope this work can bring new inspiration to few-shot learning as well as other related fields.


\keypoint{Acknowledgements.} 
This study is supported by grants from National Key R\&D Program of China (2022YFC2009903/2022YFC2009900), the National Natural Science Foundation of China (Grant No. 62122018, No. 62020106008, No. 61772116, No. 61872064), Fok Ying-Tong Education Foundation(171106), and SongShan Laboratory YYJC012022019.

{\small
\bibliographystyle{ieee_fullname}
\bibliography{egbib2}
}

\end{document}